%% file: main.tex
\documentclass{article}

\usepackage{caption}
\usepackage{multicol}
\usepackage{graphicx}
\usepackage{todonotes}
\usepackage{amsmath}
\usepackage{hyperref}
\usepackage{cleveref}
\usepackage{svg}   
\usepackage{subcaption}
\usepackage{booktabs}

 \usepackage[preprint]{corl_2026} 
   \hypersetup{pdftitle={An Adversarial Multi-Agent Pursuit Game for Locomotor Agility Pretraining},
               pdfauthor={Nathan A. C. Hogg, Matt Hyatt, Neil Klingensmith, George K. Thiruvathukal}}

\title{An Adversarial Multi-Agent Pursuit Game \\
for Locomotor Agility Pretraining}

\author{
  Nathan A. C. Hogg \quad Matt Hyatt \quad Neil Klingensmith \quad George K. Thiruvathukal \\
  Department of Computer Science, Loyola University Chicago \\
  \texttt{\{nhogg, mhyatt, nklingensmith, gthiruvathukal\}@luc.edu}
}

\begin{document}
\maketitle


\begin{abstract}
    Agile humanoid locomotion requires controllers that can adapt to rapidly changing motion demands while remaining stable and robust. 
    Recent locomotion controllers are typically designed with either a dense reward that is carefully shaped to encourage steps and penalize falling, or by using a reference tracking reward which teaches the controller to follow predetermined motions. 
    Once a behavior that maximizes the reward is discovered, the controller ceases to learn. While generally stable, reference-tracking methods focus on motor control separately from motion generation. 
    Self-play has been proven as an effective technique for superhuman performance on several multiagent games like Go, Starcraft, hide-and-seek, and capture-the-flag. We train neural networks to play the game of multiagent tag in a physics simulator to study 
    if new motion generation and control strategies can emerge through sparse feedback and competitive self-play. 
    
    We present \textbf{MJTag}, a task suite for role-conditioned locomotion during competitive pursuit-evasion. 
    Our model is a role-conditioned policy for a low-level locomotion controller and optional planar velocity commands.
MJTag's league-based-play with sparse feedback enables a locomotion controller to adopt new strategies, including dodging, sprinting, and tackling.
\end{abstract}



\input{sec/intro}
\input{sec/method}

\input{sec/results}

\input{sec/conclusion}

\clearpage
\acknowledgments{
This material is based upon work supported by the Air Force Office of Scientific Research under award number FA9550-23-F-0014 in the amount of \$153,600.

This research was supported in part by a Research Support Grant administered through the Office of Research Services at Loyola University Chicago in the amount of \$5000.
}


\bibliography{refs}  

\input{sec/A_appendix}

\end{document}

%% file: sec/intro.tex
\input{qualitative/teaser}

\section{Introduction}
Agile robotic locomotion is essential for robots to be widely useful in real-world problems such as delivery, exploration, search and rescue, and healthcare. 
Current approaches often rely on supervised imitation of reference trajectories and reinforcement learning (RL) for high frequency control.
While imitation learning provides stable training from prior demonstrations~\cite{peng2018deepmimic} , it depends on high-quality motion data. Such data is limited in scale, and it is highly dependent on the capture environment~\cite{xu_parc_2025}. Often, motion capture must be post-processed to become physically plausible. A generative locomotion controller is limited to move only in ways that fall inside the training distribution.

In contrast, RL allows agents to learn from experience but often stagnates once a reward is maximized. Standard curriculum learning is frequently used to make the environment more difficult but the curriculum is seldom aware of the capabilities of the policy leading to impossible or trivial tasks \cite{kim2025high}. This eventually dilutes the quality of training data. Advanced environment tuning results in a difficult high-dimensional optimization problem~\cite{xie2020allstepscurriculumdrivenlearningstepping}.

We hypothesize that the shortcomings of locomotor RL and motion generation are not due to any poor quality of locomotor skills but rather due to the eventual plateau of behavior when maximizing the reward function under the training environment. Additionally, the manual effort required to tune both the environment difficulty and reward shape are undesirable traits of RL. 

Neither supervised learning nor dense RL offer a simple recipe for policy improvement that is resistant to plateau. This differs from RL on Atari-style games or manipulation tasks, where the reward being maximized is itself the desired behavior~\cite{mnih2013playingatarideepreinforcement}. Conversely, terrain traversal is open-ended and any dense reward imposes a bias that is not guaranteed to be aligned with locomotion. 

In competitive games, self-play is often used to enforce an adaptive curriculum. 
A weak model can begin to learn by dominating a weak opponent, and successive training rounds will prepare the policy to learn counter-strategies in order to receive a scalar reward~\cite{alphastar,silver2017masteringgo,chen2019visualhideseek,team2021open}. In this case, even though the earned reward is unchanged, the model still adapts since a skilled opponent will force the player into undesirable state-space which it will need to overcome. 

Recently, self-play was used as a strategy in drone pursuit~\cite{tang2026agilepeautonomousuavpursuitevasion}, demonstrating impressive low-level flight control at high frequency. 
Previously, tag has been used in quadrupedal motion~\cite{tang2020learningagilelocomotionadversarial} although their method has some notable differences to ours, with the primary difference being that their evader is a 2D dot coordinate rather than another identical robot entity.
This makes their training protocol more similar to goal following than self-play. 
Moreover, they report that their model cannot learn from RL and they opt for evolutionary algorithms. Han et al. also train quadrupeds using a game of tag~\cite{Han_2024_tencent_lifelike_agile}, but they freeze the locomotion controller and train a goal-proposal policy. This means that the game does not teach locomotor strategy. Instead, it teaches the high-level selection of planar coordinate goals. Similar to us, Han et al. are inspired by the parkour game World Chase Tag. 
Finally, Prioritized Fictitious Self-Play (PFSP)~\cite{alphastar} has been used in the learning of character controllers \cite{zhu2023NCP} for boxing matches.

To address limitations in environment difficulty and reward modeling, we propose competitive tag as a surrogate objective that is well aligned with terrain-traversal. We hypothesize that the policy should continue to improve even on simple terrains since PFSP enforces adaptive difficulty.

Our contributions include (1) the development of a multiagent competitive league environment for humanoids, 
(2) a data sampling strategy to balance role-difficulty imbalances, and (3) a comparative analysis of state exploration under self-play and dense reward shaping.
We demonstrate that competitive league-based play with sparse feedback can train a locomotion controller to develop strategic skills like lunging, tackling and ducking. 

%% file: qualitative/teaser.tex
\begin{figure}[hb]
  \centering
  \includegraphics[width=0.331\linewidth]{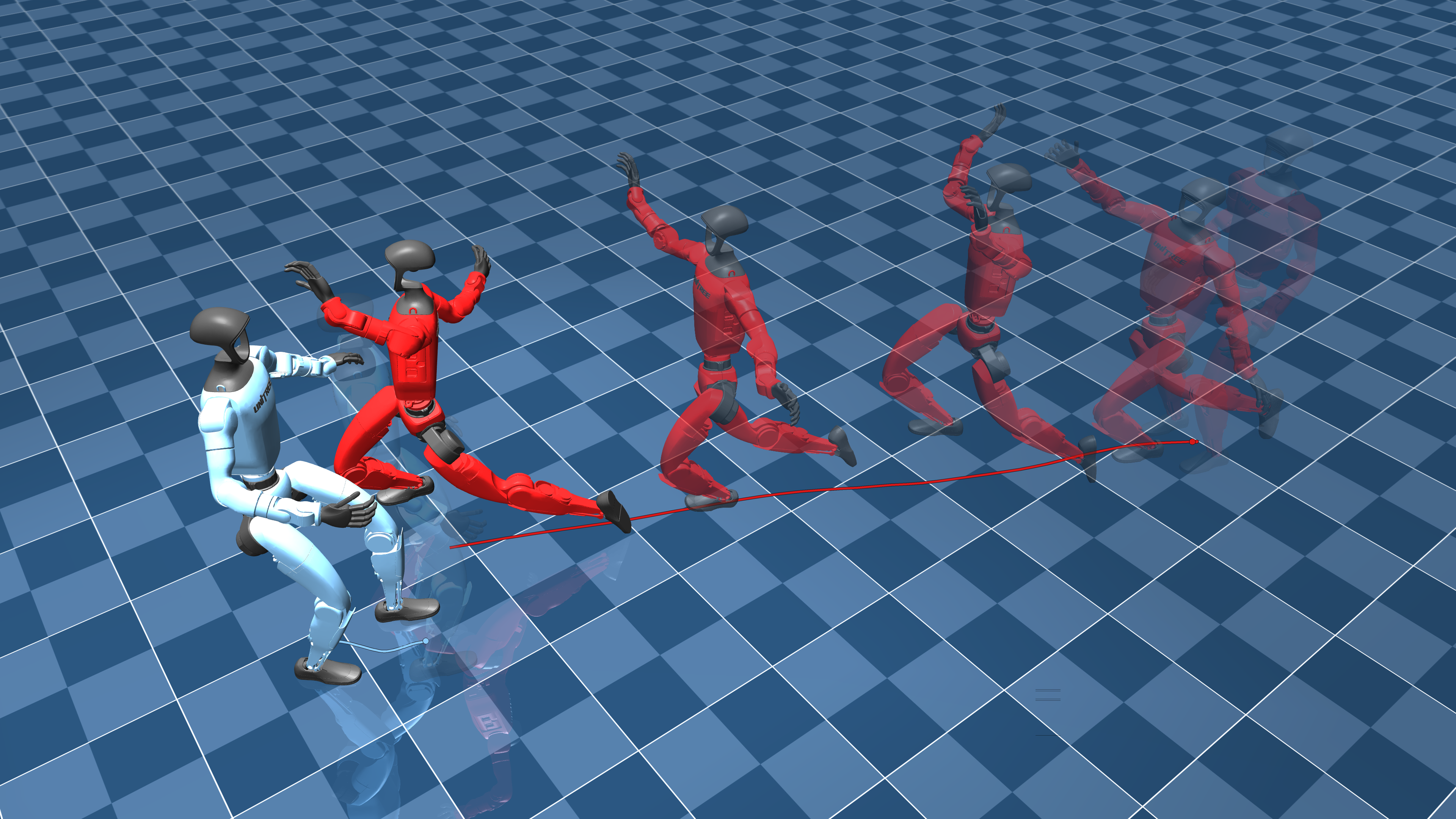}\hfill
  \includegraphics[width=0.331\linewidth]{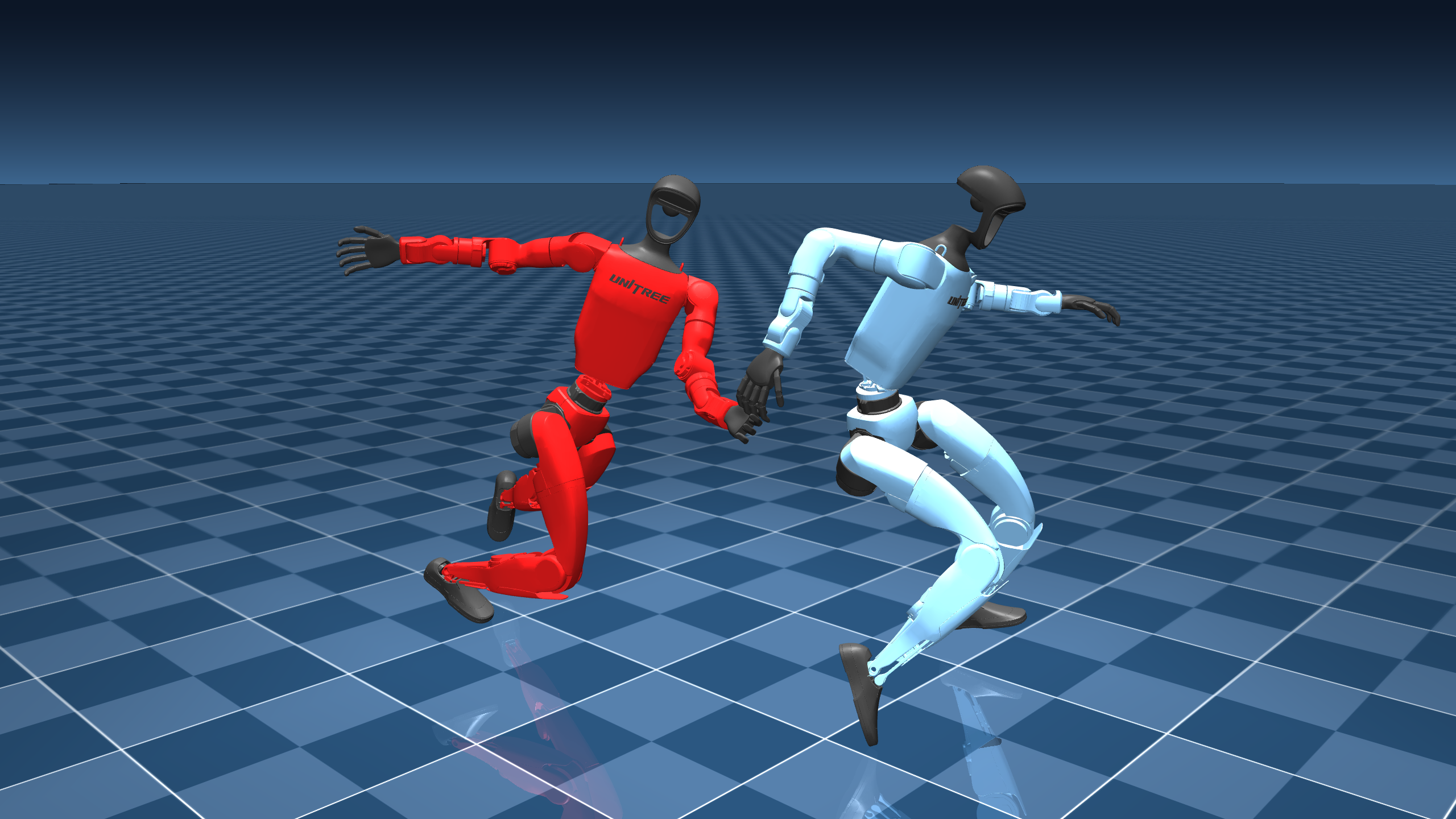}\hfill
  \includegraphics[width=0.331\linewidth]{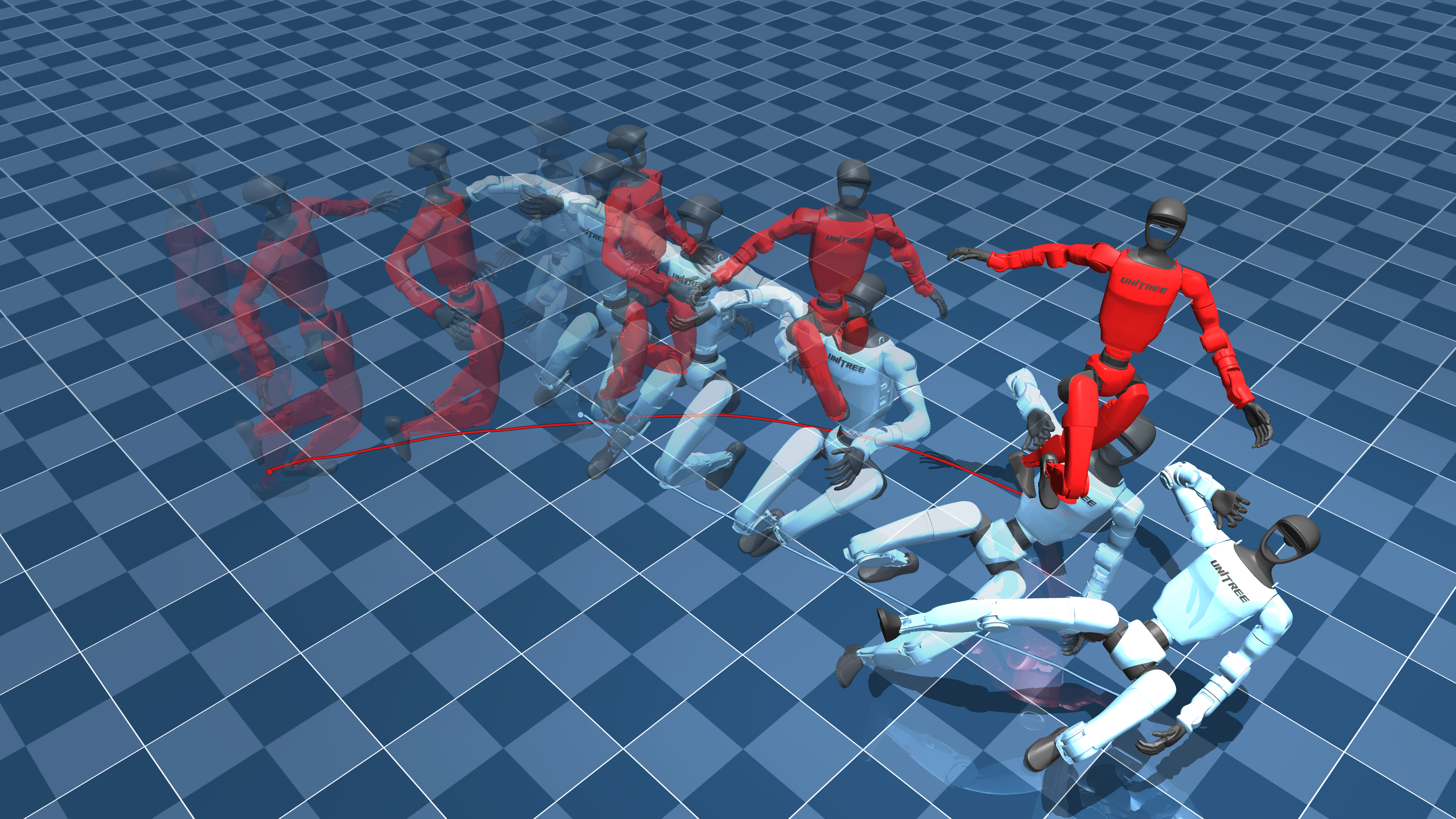}

  \smallskip

  \includegraphics[width=0.248\linewidth]{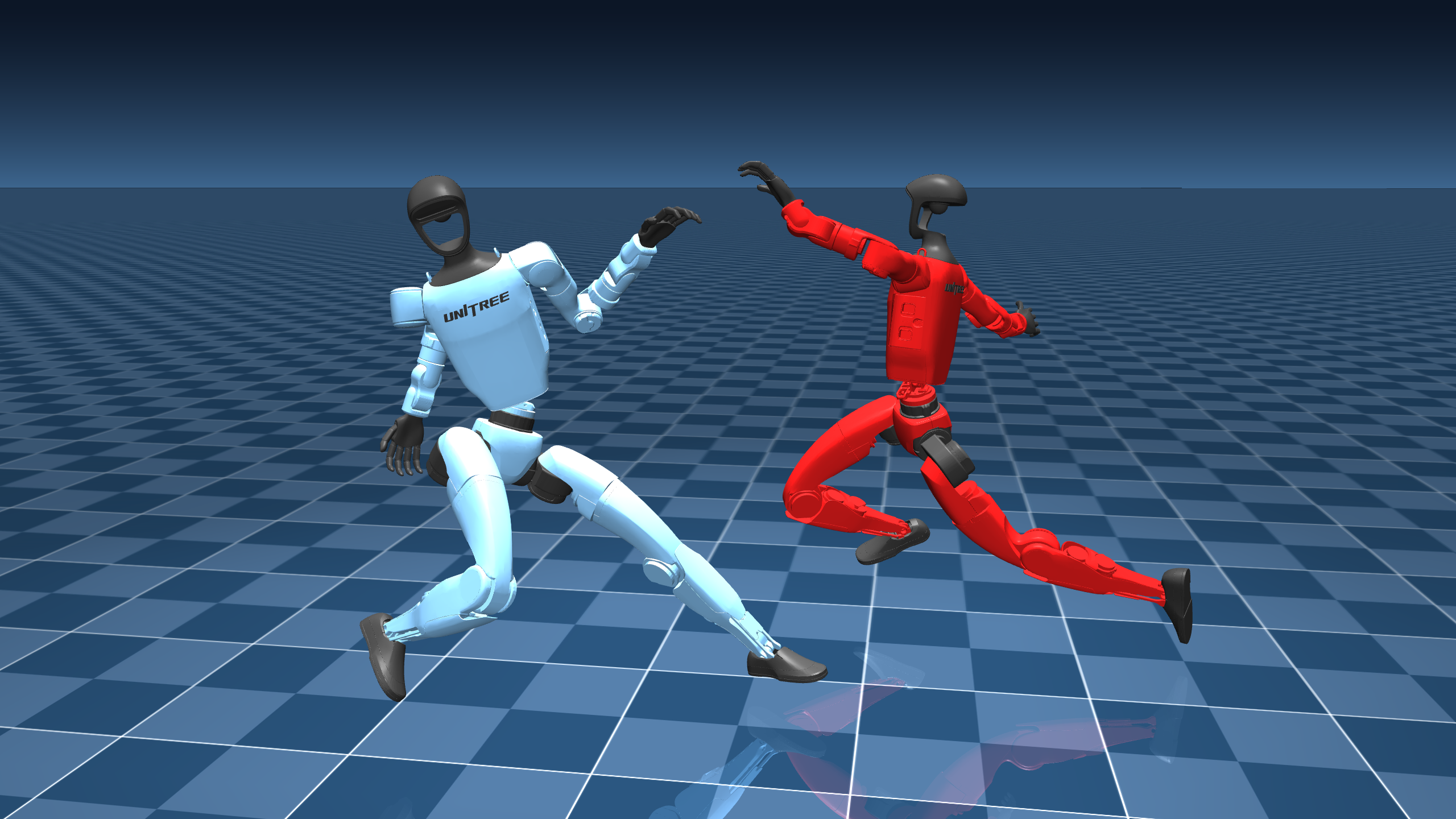}\hfill
  \includegraphics[width=0.248\linewidth]{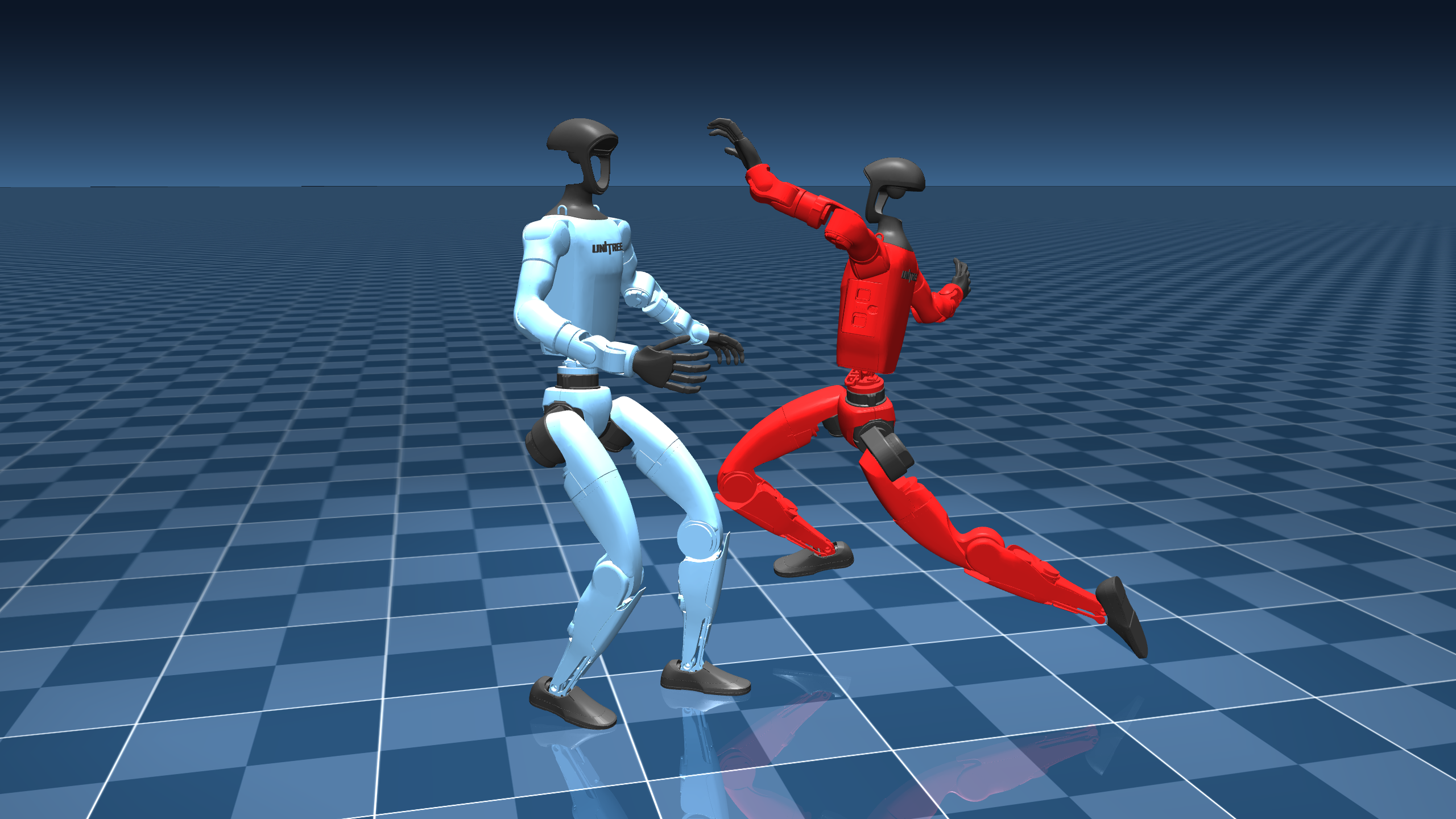}\hfill
  \includegraphics[width=0.248\linewidth]{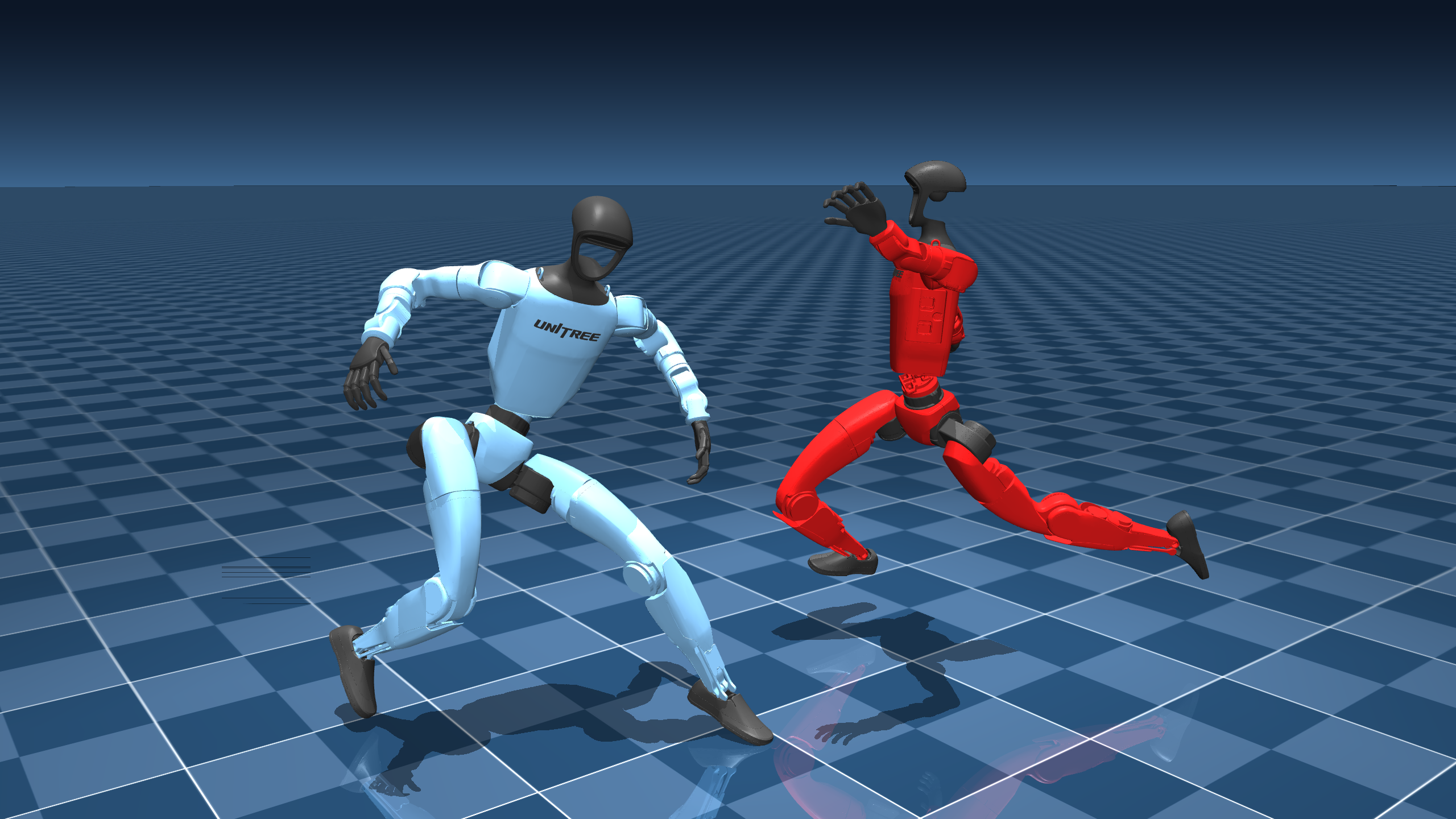}\hfill
  \includegraphics[width=0.248\linewidth]{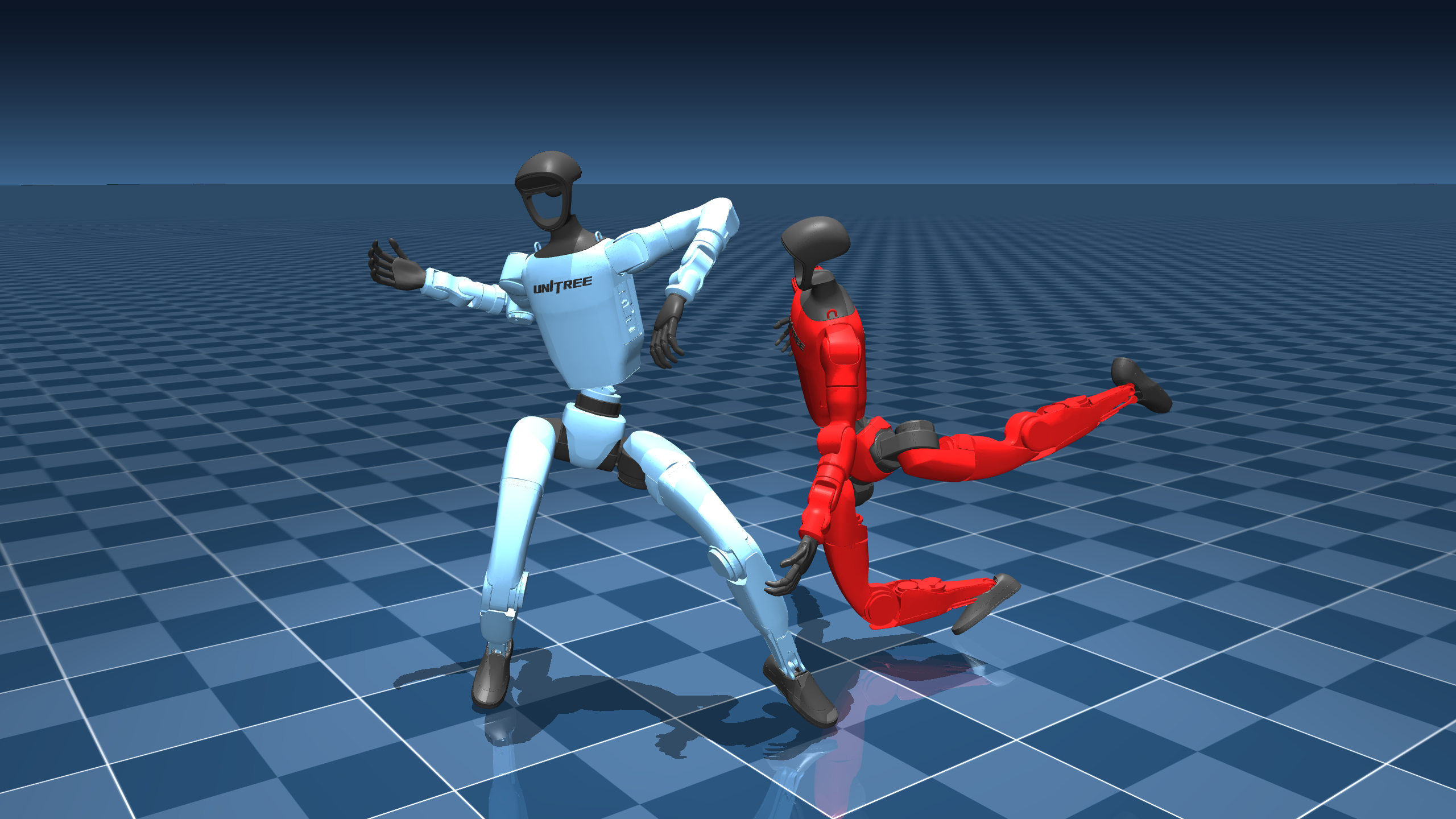}

  \smallskip

  \caption{Simulated humanoid agents play tag: demonstrating lunging, diving, dodging and chasing.}
  \label{fig:grid}
\end{figure}

%% file: sec/method.tex
\section{Multi-Agent Pursuit}
\label{sec:method}

This section outlines environment design, data sampling, task specification, and rewards. For all experiments, we train an MLP using PPO~\cite{schulman2017proximalpolicyoptimizationalgorithms}. 
Contrary to previous work~\cite{tang2020learningagilelocomotionadversarial}, our model is conditioned on its role, which allows the same model to perform pursuit and evasion.

\begin{figure}[tb]
  \centering
  \includegraphics[width=\linewidth]{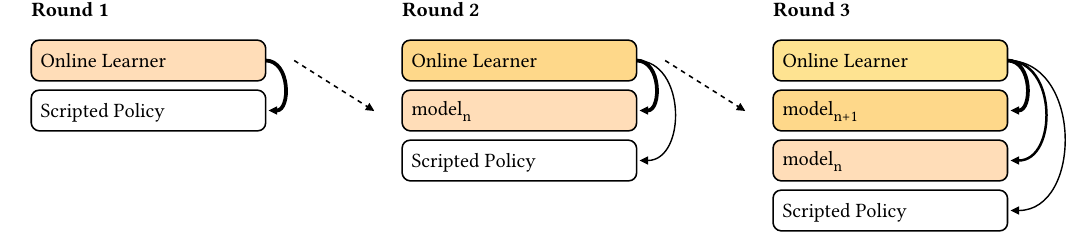}
  \caption{PFSP league training. }
  \label{fig:bracket}
\end{figure}

The environment uses MuJoCo Warp via the \texttt{mjlab}~\cite{zakka2026mjlablightweightframeworkgpuaccelerated} library. 
Physics runs at 200\,Hz with 4 substeps and policy control at 50\,Hz.
The environment contains two Unitree G1 humanoids. 
Each humanoid has 29 actuators, for 58 total actuators in each simulated environment.

\textbf{Opponent Sampling:} We select opponents with Prioritized Fictitious Self-Play (PFSP)~\cite{alphastar,tang2026agilepeautonomousuavpursuitevasion}. Every 2,500 PPO epochs we add a frozen copy of the online model to a first-in, first-out league of 32 checkpoints, which also always contains a scripted policy that outputs zero actions, and serves as a passive baseline; Fig.~\ref{fig:bracket}. Following \cite{alphastar}, each opponent is sampled with weight $f(p)=(1-p)^2$, where $p$ is the learner's empirical win rate against it, so opponent difficulty is measured from outcomes rather than assumed from training age.

\textbf{Role Sampling:} We use an online adaptive sampling mechanism to address the instability associated with role-difficulty imbalance: it is overwhelmingly easier for the models to play as the pursuer rather than the evader. Naively, this leads to training collapse and a lack of meaningful policy feedback. This imbalance is less present in other multiagent games where the objective is the same for both agents. 
We initialize the role sampler at $\rho=0.5$ at the beginning of training and adjust the sampling weight after each round. If the evader win-rate (timeout rate) is less than 50\%, we increase the role sampling weight by $\delta_\rho=0.05$ and reduce the learner evader role probability if at least 50\% of data rollouts time out. This gives the policy more evasion experience and keeps play competitive. Ablations show that without this sampling mechanism training is unstable.
We present a side-by-side comparison that demonstrates how role sampling improves the balance of skill between the pursuer and evader in Appendix~\ref{sec:appendix}.


\textbf{Rewards:} The policy receives a semi-sparse reward with weak dense shaping. 
Falls and arena exits terminate the episode and apply a -10 penalty to the offending agent. The environment considers an agent fallen if torso tilt is greater than $70^\circ$ or pelvis height is below $0.5 \mathrm{m}$, while leaving the arena triggers the out-of-bounds condition. \Cref{eg:reward} shows this formulation:

\begin{equation}
\label{eg:reward}
r^{i} = w_{\mathrm{fall}} * r^{i}_{\mathrm{fall}} + r^i_{\mathrm{lim}} +  \sigma_i*\,\bigl(w_{\mathrm{tag}}*r_{\mathrm{tag}} + r_{\mathrm{gap}}\bigr),
\qquad
\sigma_i =
\begin{cases}
+1 & i = \text{pursuer}\\
-1 & i = \text{evader}
\end{cases}\end{equation}

Here $r^i_{\text{fall}}{=}{-1}$ if agent $i$ falls or leaves the arena, ending the episode, and $0$ otherwise; $r^i_{\text{lim}}$ is a small joint-limit penalty; $r_{\text{tag}}{=}1$ when the agents come within $0.5$\,m in the $(x,y)$ plane, which also ends the episode; and $r_{\text{gap}}{=}{-}w_{\text{gap}}(d_t{-}d_{t-1})$, where $d_t$ is their planar distance. $\sigma_i$ ensures that role-based competition terms are treated as rewards or penalties for the respective roles. We use $w_{\text{fall}}{=}10$, $w_{\text{tag}}{=}5$, and $w_{\text{gap}}{=}0.001$.

%% file: sec/results.tex
\section{Experimental Results}
\label{sec:result}

\begin{figure}[t]
  \begin{minipage}[t]{0.5\linewidth} 
    \centering
    \includegraphics[width=\linewidth]{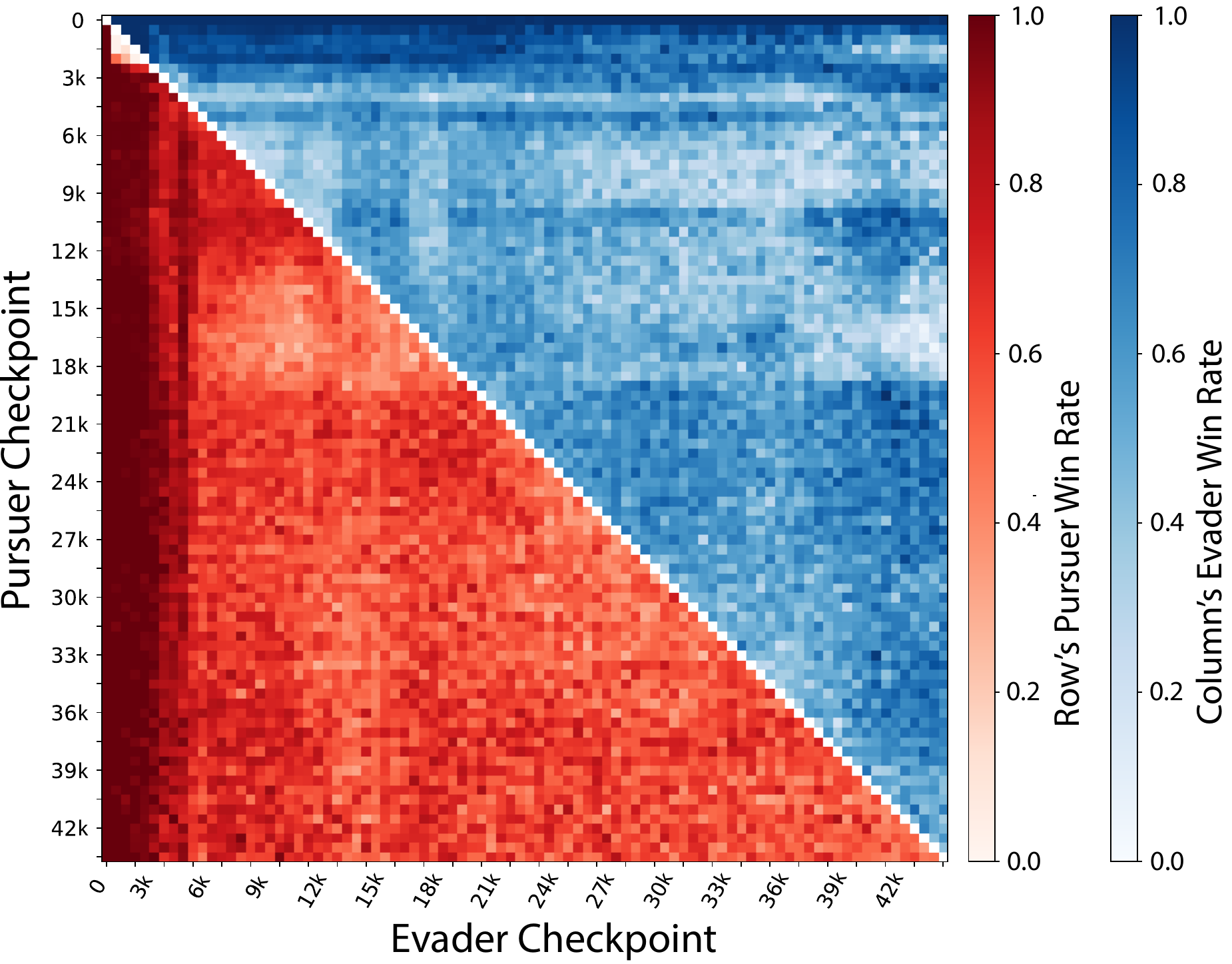}
    \caption{Tag win-rate between league matches: $x$-axis is the evader checkpoint, $y$-axis is the pursuer checkpoint. 
     }
    \label{fig:league}
  \end{minipage}
  \hspace{0.1cm}
  \begin{minipage}[t]{0.5\linewidth} 
    \includegraphics[width=\linewidth]{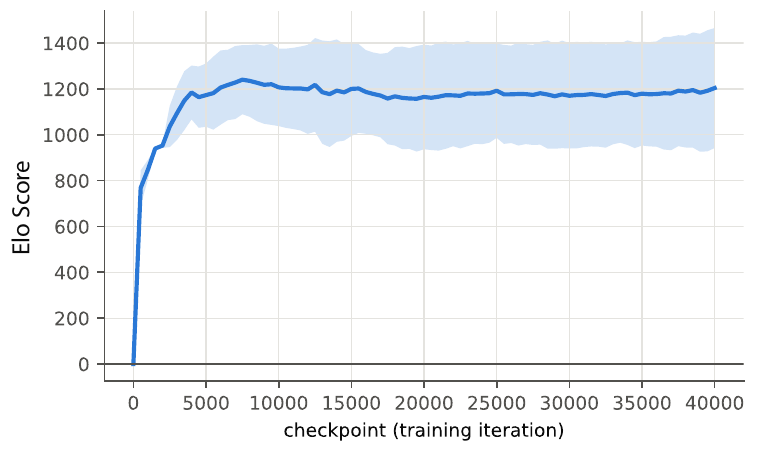}
    \caption{
        Elo score of the online learner over 40,000 rounds of play (3.9 Billion env steps).
    }
    \label{fig:elo-step}
  \end{minipage}
\end{figure}

In our evaluation, the locomotion controller adopts interesting behaviors as it attempts to win the game of tag against competitors in the league.
We also observe that Elo rises sharply in early training, and the league continues to produce new strategies after the rating levels off.

\Cref{fig:league} shows the color-coded win rate of the locomotion controller against other competitors in the league.
In the topmost row, the results show 100\% success, indicating that all subsequent models learn to win against the untrained pursuer. The dark blue color indicates evader success. Similarly, all pursuers win against the first evader (red). 
In subsequent rounds, the online learner competes against other checkpointed competitors from the league.
At $y=19000$, the map becomes darker red, indicating that the pursuer learns a new strategy that improves wins against previous evaders. Similarly from $x=[12000,16000]$ and $[39000,43000]$ the evader improves against previous pursuers.

We evaluate progress with Elo ratings~\cite{arpad_elo1961new, tang2025eloratingreliablestudy}, computed purely from game outcomes with no human judgment. Each saved checkpoint plays 4096 deterministic games, one episode per parallel world from a fixed reset seed, split evenly among all earlier checkpoints, with roles assigned uniformly at random in each game (win $=1$, draw $=0.5$, loss $=0$). A pair of checkpoints thus plays $4096/k$ games, where $k$ is the number of earlier checkpoints. We fit a Bradley--Terry model to all outcomes jointly, adding one drawn pseudo-game per pair so that ratings stay finite under a clean sweep. Maximum-likelihood ratings are reported on the Elo scale ($400/\ln 10$ per logit), anchored at the untrained policy (0 Elo); lines show the mean across environments and the shaded band spans minimum to maximum.

\textbf{Learned Agility: }
Figure~\ref{fig:elo-step} shows Elo rising sharply over the first $\sim$5{,}000 iterations and then holding near 1200. Because ratings are fit only against the league's own checkpoints, a flat curve means later checkpoints are not uniformly stronger than earlier ones; it does not mean behavior has stopped changing. Figure~\ref{fig:league} shows distinct strategy shifts after this point: the pursuer at $y{=}19$k and the evader at $x{=}39$--$43$k each beat earlier opponents of the other role. We read this as the two roles trading advantages rather than settling into one fixed behavior. Qualitatively, the controller learns to lunge, dodge, and dive. We do not develop a formal agility measure; Elo measures strength relative to prior checkpoints only.

%% file: sec/conclusion.tex
\section{Conclusion}
\label{sec:conclusion}

In our experiments, the controllers frequently exhibit unusual motions. Because the policy is not rewarded for tracking reference motions or staying near a behavioral prior~\cite{he2025hoverversatileneuralwholebody}, some of these may be effective exploits of the simulator's contact model rather than poor play.
Areas for future work include adversarial terrain traversal subject to natural motion penalties and using self-play to explore efficient motions which still fall in-distribution of human motion generators. 

MJTag applies competitive league-based training for RL in robotics, which incrementally incentivizes controllers to adopt new behaviors to win in a tournament. We show that tournament-style training continues to produce new locomotor strategies, such as lunging, dodging, and diving, from a sparse game outcome. Limitations are discussed in Appendix~\ref{app:limitations}.

%% file: sec/A_appendix.tex
\newpage

\appendix

\section{Appendix}

\label{sec:appendix}

Here we present additional evaluation to demonstrate the performance of adversarial play.

\begin{figure}[htbp]
  \centering
  \begin{subfigure}[t]{0.32\linewidth}
    \centering
    \includegraphics[width=\linewidth]{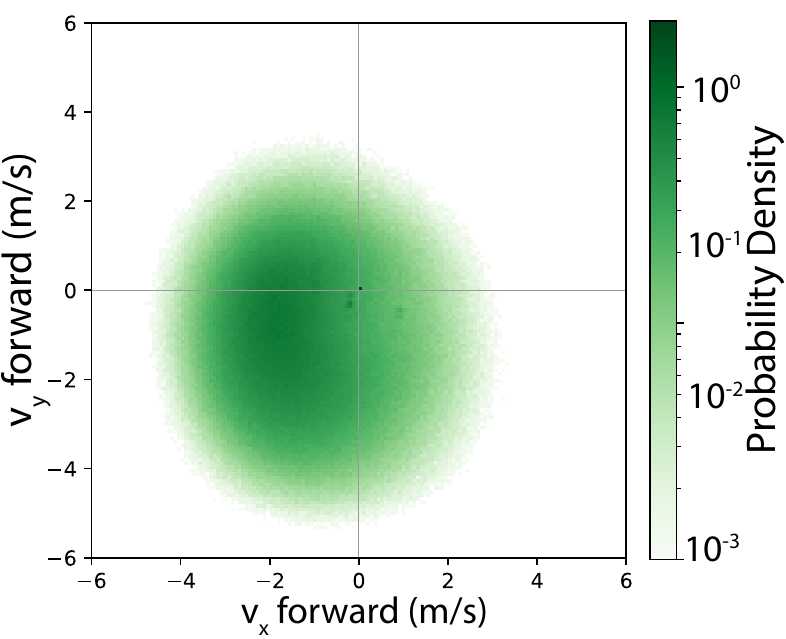}
    \caption{Dense Goal-following}
    \label{fig:untrainedvelocity}
  \end{subfigure}%
  \hfill
  \begin{subfigure}[t]{0.32\linewidth}
    \centering
    \includegraphics[width=\linewidth]{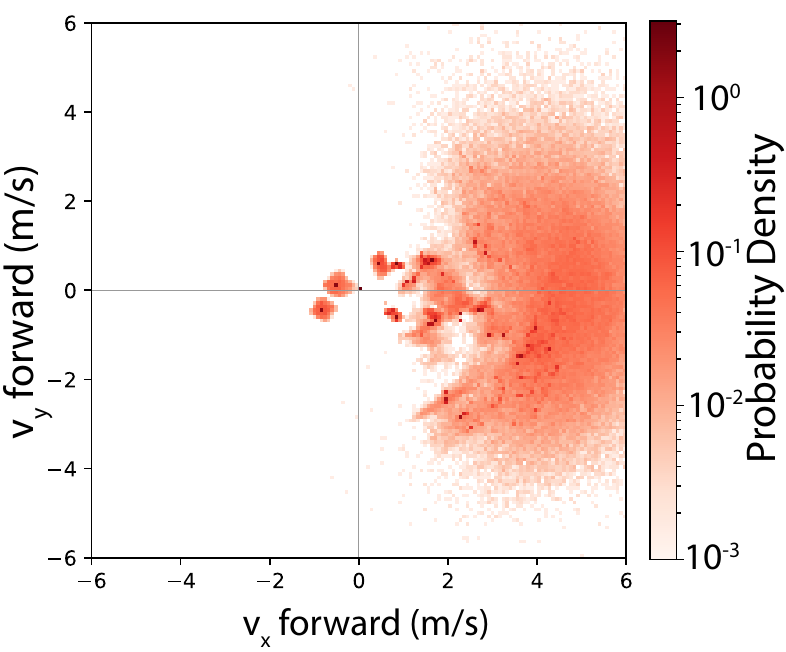}
    \caption{Pursuer}
    \label{fig:trainedvelocitypursuer}
  \end{subfigure}%
  \hfill
  \begin{subfigure}[t]{0.32\linewidth}
    \centering
    \includegraphics[width=\linewidth]{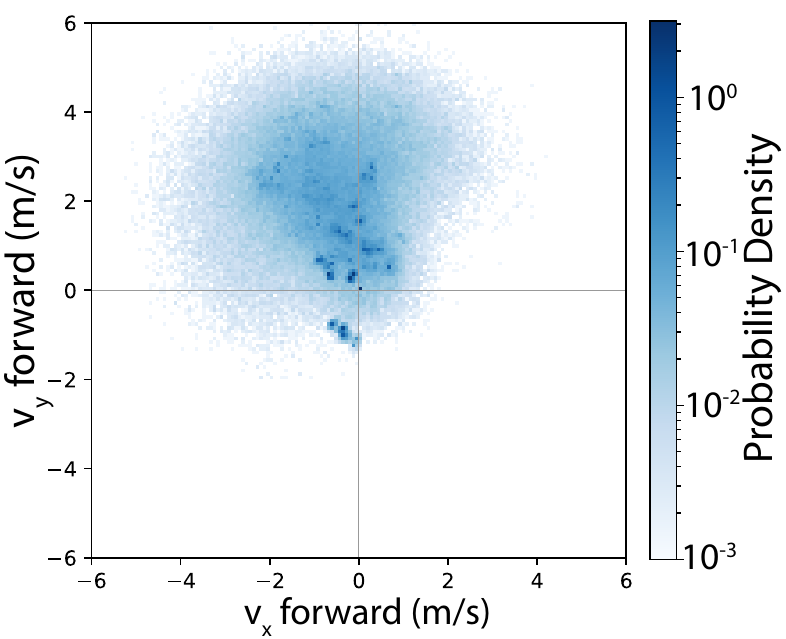}
    \caption{Evader}
    \label{fig:trainedvelocityevader}
  \end{subfigure}
  \caption{Probability density of the velocity vector
    (a)~for a naive dense reward without league play, and for the
    (b)~pursuer and (c)~evader after 40,000 rounds of tag.}
  \label{fig:velocity}
\end{figure}

\textbf{Learned Movement Characteristics: }
\Cref{fig:untrainedvelocity} shows the probability density of planar velocity of a standard model trained with goal-tracking dense reward. This model did not participate in league play.
Overall, the model does not initially exhibit a strong preference for moving in any particular direction.
Figures \ref{fig:trainedvelocitypursuer} and \ref{fig:trainedvelocityevader} show the probability density function of the velocity vectors of the pursuer and the evader after participating in 43,000 rounds of league play.
After training, the controller clearly has learned a strategy that is different as a result of conditioning the model on the pursuer role and evader role. Moreover, the policy continually seeks to push the boundary of its capability and changes its exploration strategy over the course of training.

\begin{figure}[htbp]
  \begin{minipage}[b]{0.5\linewidth}
    \centering
    \includegraphics[width=\linewidth]{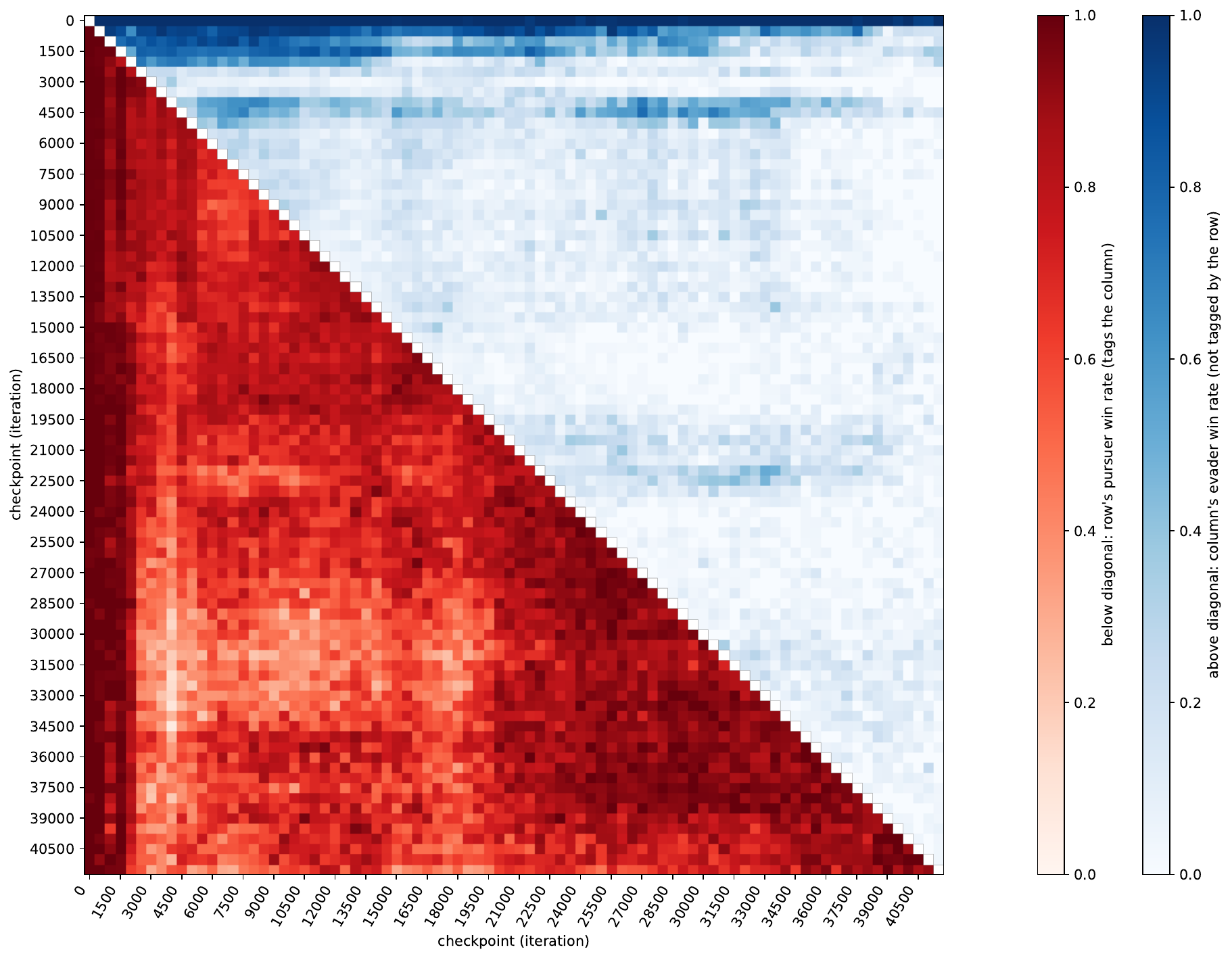}
    \caption{Win rate \textbf{without} role sampling.}
    \label{fig:norolesampling}
  \end{minipage}
  \hspace{0.5cm}
  \begin{minipage}[b]{0.5\linewidth}
    \centering
    \includegraphics[width=\linewidth]{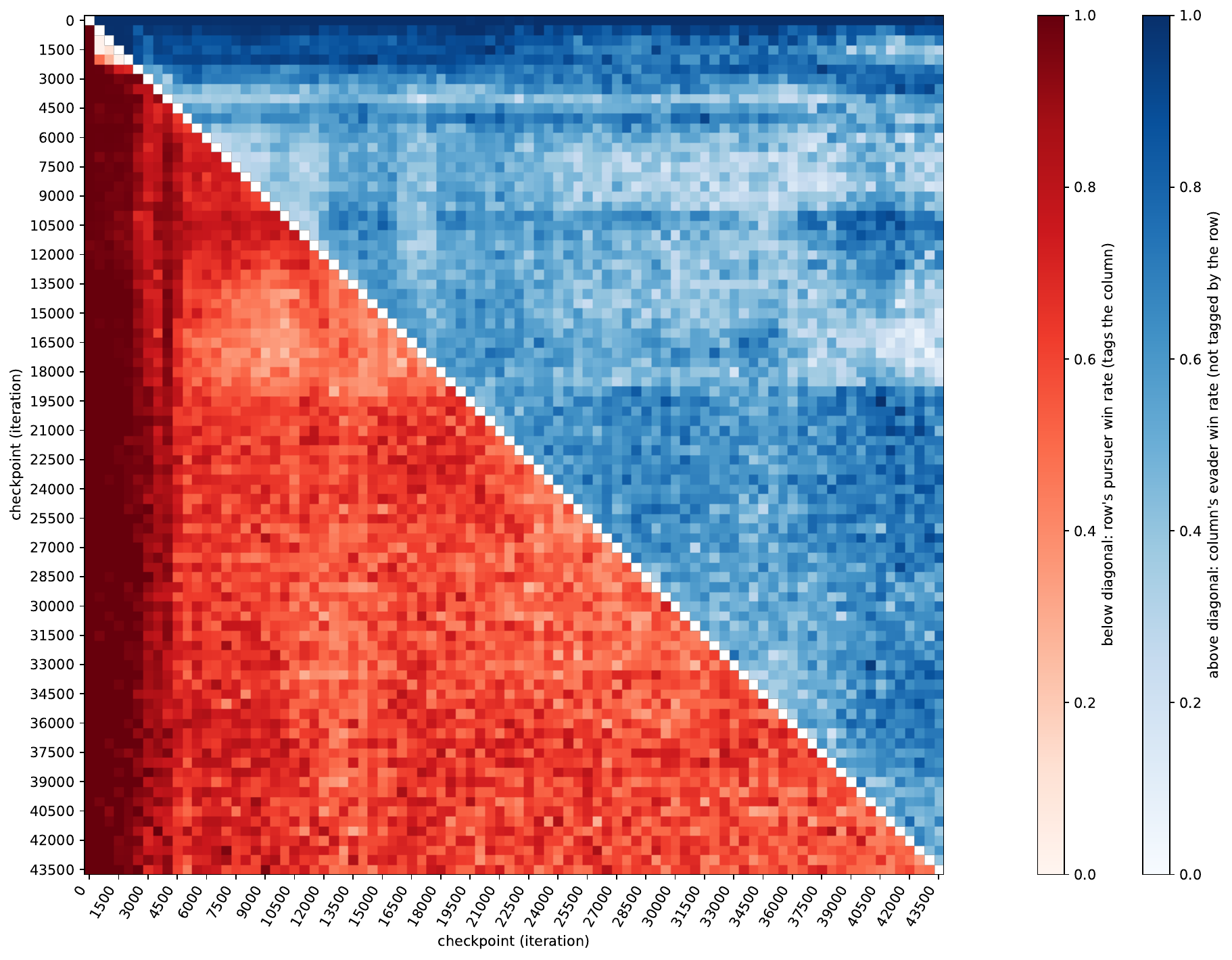}
    \caption{Win rate \textbf{with} role sampling.}
    \label{fig:rolesampling}
  \end{minipage}
\end{figure}

\textbf{Role Sampling: }
Figures \ref{fig:norolesampling} and \ref{fig:rolesampling} demonstrate how role sampling improves the balance of performance between pursuer and evader.
Without role sampling (\Cref{fig:norolesampling}), the pursuer's skill increases much faster than that of the evader, and after several hundred rounds wins consistently.

\section{Limitations}\label{app:limitations}

\paragraph{Scope of evaluation.} All results are in simulation on flat terrain, from a single training seed. We measure progress with Elo relative to the league's own checkpoints and do not yet compare against imitation-learning or dense-reward RL baselines on a common metric, nor do we define a formal measure of agility.

\paragraph{Task definition.} A tag is defined by planar proximity (0.5\,m) rather than physical contact. This may make the task easier than true tag and limits the role of deliberate contact skills such as tackling.

\paragraph{Scope of learned skills.} Self-play rewards only skills that help win the game. Behaviors outside it, such as stair climbing, carrying loads, or a natural gait style, still require reward shaping or imitation. Because the policy is not regularized toward reference motions, it frequently produces unusual motions; some may be effective exploits of the simulator's contact model rather than poor play, and we have not yet separated the two.

\paragraph{Sim-to-real transfer.} We do not deploy on hardware. Transfer would face safety risks from two robots colliding at speed, hardware wear from repeated falls, contact and friction mismatch during lunges and dives, and the need to estimate the opponent's state from onboard sensing.

  Code and the MJTag environment are available at \url{https://github.com/LUC-Robotics-Lab/tag}

%% file: refs.bib
@article{
        zhu2023NCP,
        author = {Zhu, Qingxu and Zhang, He and Lan, Mengting and Han, Lei},
        title = {Neural Categorical Priors for Physics-Based Character Control},
        year = {2023},
        issue_date = {December 2023},
        publisher = {Association for Computing Machinery},
        address = {New York, NY, USA},
        volume = {42},
        number = {6},
        issn = {0730-0301},
        url = {https://doi.org/10.1145/3618397},
        doi = {10.1145/3618397},
        journal = {ACM Trans. Graph.},
        month = {dec},
        articleno = {178},
        numpages = {16}
}

@article{Han_2024_tencent_lifelike_agile,
   title={Lifelike agility and play in quadrupedal robots using reinforcement learning and generative pre-trained models},
   volume={6},
   ISSN={2522-5839},
   url={http://dx.doi.org/10.1038/s42256-024-00861-3},
   DOI={10.1038/s42256-024-00861-3},
   number={7},
   journal={Nature Machine Intelligence},
   publisher={Springer Science and Business Media LLC},
   author={Han, Lei and Zhu, Qingxu and Sheng, Jiapeng and Zhang, Chong and Li, Tingguang and Zhang, Yizheng and Zhang, He and Liu, Yuzhen and Zhou, Cheng and Zhao, Rui and Li, Jie and Zhang, Yufeng and Wang, Rui and Chi, Wanchao and Li, Xiong and Zhu, Yonghui and Xiang, Lingzhu and Teng, Xiao and Zhang, Zhengyou},
   year={2024},
   month=jul, pages={787–798} }

@misc{tang2020learningagilelocomotionadversarial,
      title={Learning Agile Locomotion via Adversarial Training}, 
      author={Yujin Tang and Jie Tan and Tatsuya Harada},
      year={2020},
      eprint={2008.00603},
      archivePrefix={arXiv},
      primaryClass={cs.RO},
      url={https://arxiv.org/abs/2008.00603}, 
}

@misc{chen2019visualhideseek,
      title={Visual Hide and Seek}, 
      author={Boyuan Chen and Shuran Song and Hod Lipson and Carl Vondrick},
      year={2019},
      eprint={1910.07882},
      archivePrefix={arXiv},
      primaryClass={cs.AI},
      url={https://arxiv.org/abs/1910.07882}, 
}

@misc{mnih2013playingatarideepreinforcement,
      title={Playing Atari with Deep Reinforcement Learning}, 
      author={Volodymyr Mnih and Koray Kavukcuoglu and David Silver and Alex Graves and Ioannis Antonoglou and Daan Wierstra and Martin Riedmiller},
      year={2013},
      eprint={1312.5602},
      archivePrefix={arXiv},
      primaryClass={cs.LG},
      url={https://arxiv.org/abs/1312.5602}, 
}

@misc{schulman2017proximalpolicyoptimizationalgorithms,
      title={Proximal Policy Optimization Algorithms}, 
      author={John Schulman and Filip Wolski and Prafulla Dhariwal and Alec Radford and Oleg Klimov},
      year={2017},
      eprint={1707.06347},
      archivePrefix={arXiv},
      primaryClass={cs.LG},
      url={https://arxiv.org/abs/1707.06347}, 
}

@misc{zakka2026mjlablightweightframeworkgpuaccelerated,
  title={mjlab: A Lightweight Framework for GPU-Accelerated Robot Learning},
  author={Kevin Zakka and Qiayuan Liao and Brent Yi and Louis Le Lay and Koushil Sreenath and Pieter Abbeel},
  year={2026},
  eprint={2601.22074},
  archivePrefix={arXiv},
  primaryClass={cs.RO},
  url={https://arxiv.org/abs/2601.22074},
}

@article{arpad_elo1961new,
  author  = {Elo, Arpad E.},
  title   = {The New {USCF} Rating System},
  journal = {Chess Life},
  volume  = {16},
  pages   = {160--161},
  year    = {1961}
}

@misc{tang2025eloratingreliablestudy,
      title={Is Elo Rating Reliable? A Study Under Model Misspecification}, 
      author={Shange Tang and Yuanhao Wang and Chi Jin},
      year={2025},
      eprint={2502.10985},
      archivePrefix={arXiv},
      primaryClass={cs.LG},
      url={https://arxiv.org/abs/2502.10985}, 
}

@misc{tang2026agilepeautonomousuavpursuitevasion,
      title={AgilePE: Autonomous UAV Pursuit-Evasion via Self-Play Reinforcement Learning}, 
      author={Wenhao Tang and Tianyang Chen and Zhejun Cui and Boyuan An and Jiayu Chen and Ruize Zhang and Huidong Liu and Tianyue Wu and Qingmin Liao and Fei Gao and Yu Wang and Chao Yu},
      year={2026},
      eprint={2608.14135},
      archivePrefix={arXiv},
      primaryClass={cs.RO},
      url={https://arxiv.org/abs/2608.14135}, 
}

@article{xie2020allstepscurriculumdrivenlearningstepping,
      title={ALLSTEPS: Curriculum-driven Learning of Stepping Stone Skills}, 
      author={Zhaoming Xie and Hung Yu Ling and Nam Hee Kim and Michiel van de Panne},
      year={2020},
      eprint={2005.04323},
      archivePrefix={arXiv},
      primaryClass={cs.GR},
    journal = {Computer Graphics Forum},
    volume = {39},
    number = {8},
    pages = {213-224},
    doi = {10.1111/cgf.14115},
      url={https://arxiv.org/abs/2005.04323}
}

@misc{xu_parc_2025,
	title = {{PARC}: {Physics}-based {Augmentation} with {Reinforcement} {Learning} for {Character} {Controllers}},
	shorttitle = {{PARC}},
	url = {http://arxiv.org/abs/2505.04002},
	doi = {10.1145/3721238.3730616},
	urldate = {2025-07-16},
	author = {Xu, Michael and Shi, Yi and Yin, KangKang and Peng, Xue Bin},
	month = may,
	year = {2025},
	note = {arXiv:2505.04002 [cs]},
}

@misc{team2021open,
      title={Open-Ended Learning Leads to Generally Capable Agents}, 
      author={{Open Ended Learning Team} and Adam Stooke and Anuj Mahajan and Catarina Barros and Charlie Deck and Jakob Bauer and Jakub Sygnowski and Maja Trebacz and Max Jaderberg and Michael Mathieu and Nat McAleese and Nathalie Bradley-Schmieg and Nathaniel Wong and Nicolas Porcel and Roberta Raileanu and Steph Hughes-Fitt and Valentin Dalibard and Wojciech Marian Czarnecki},
      year={2021},
      eprint={2107.12808},
      archivePrefix={arXiv},
      primaryClass={cs.LG},
      url={https://arxiv.org/abs/2107.12808}, 
}

@article{peng2018deepmimic,
  title={Deepmimic: Example-guided deep reinforcement learning of physics-based character skills},
  author={Peng, Xue Bin and Abbeel, Pieter and Levine, Sergey and Van de Panne, Michiel},
  journal={ACM Transactions On Graphics (TOG)},
  volume={37},
  number={4},
  pages={1--14},
  year={2018},
  publisher={ACM New York, NY, USA}
}

@article{silver2017masteringgo,
  title={Mastering the game of go without human knowledge},
  author={Silver, David and Schrittwieser, Julian and Simonyan, Karen and Antonoglou, Ioannis and Huang, Aja and Guez, Arthur and Hubert, Thomas and Baker, Lucas and Lai, Matthew and Bolton, Adrian and others},
  journal={nature},
  volume={550},
  number={7676},
  pages={354--359},
  year={2017},
  publisher={Nature Publishing Group UK London}
}

@article{kim2025high,
  title={High-speed control and navigation for quadrupedal robots on complex and discrete terrain},
  author={Kim, Hyeongjun and Oh, Hyunsik and Park, Jeongsoo and Kim, Yunho and Youm, Donghoon and Jung, Moonkyu and Lee, Minho and Hwangbo, Jemin},
  journal={Science Robotics},
  volume={10},
  number={102},
  pages={eads6192},
  year={2025},
  publisher={American Association for the Advancement of Science}
}

@misc{he2025hoverversatileneuralwholebody,
      title={HOVER: Versatile Neural Whole-Body Controller for Humanoid Robots}, 
      author={Tairan He and Wenli Xiao and Toru Lin and Zhengyi Luo and Zhenjia Xu and Zhenyu Jiang and Jan Kautz and Changliu Liu and Guanya Shi and Xiaolong Wang and Linxi Fan and Yuke Zhu},
      year={2025},
      eprint={2410.21229},
      archivePrefix={arXiv},
      primaryClass={cs.RO},
      url={https://arxiv.org/abs/2410.21229}, 
}

@article{alphastar,
  author  = {Vinyals, Oriol and Babuschkin, Igor and Czarnecki, Wojciech M. and Mathieu, Micha\`{e}l and Dudzik, Andrew and Chung, Junyoung and Choi, David H. and Powell, Richard and Ewalds, Timo and Georgiev, Petko and Oh, Junhyuk and Horgan, Dan and Kroiss, Manuel and Danihelka, Ivo and Huang, Aja and Sifre, Laurent and Cai, Trevor and Agapiou, John P. and Jaderberg, Max and Vezhnevets, Alexander S. and Leblond, R{\'e}mi and Pohlen, Tobias and Dalibard, Valentin and Budden, David and Sulsky, Yury and Molloy, James and Paine, Tom L. and Gulcehre, Caglar and Wang, Ziyu and Pfaff, Tobias and Wu, Yuhuai and Ring, Roman and Yogatama, Dani and W{\"u}nsch, Dario and McKinney, Katrina and Smith, Oliver and Schaul, Tom and Lillicrap, Timothy and Kavukcuoglu, Koray and Hassabis, Demis and Apps, Chris and Silver, David},
  title   = {Grandmaster level in StarCraft II using multi-agent reinforcement learning},
  journal = {Nature},
  volume  = {575},
  number  = {7782},
  pages   = {350--354},
  year    = {2019},
  doi     = {10.1038/s41586-019-1724-z}
}
